\documentclass[journal]{IEEEtran}

\usepackage{xcolor,soul,framed} 

\colorlet{shadecolor}{yellow}
\usepackage[pdftex]{graphicx}
\graphicspath{{../pdf/}{../jpeg/}}
\DeclareGraphicsExtensions{.pdf,.jpeg,.png}

\usepackage[cmex10]{amsmath}
\usepackage{array}
\usepackage{mdwmath}
\usepackage{mdwtab}
\usepackage{eqparbox}
\usepackage{url}
\usepackage{hyperref}
\usepackage{amsmath}
\usepackage{amssymb}
\usepackage{ragged2e} 
\usepackage{booktabs,makecell, multirow, tabularx}
\usepackage[ruled,linesnumbered]{algorithm2e}
\usepackage[numbers]{natbib}
\SetKwComment{Comment}{// }{}

\def\BibTeX{{\rm B\kern-.05em{\sc i\kern-.025em b}\kern-.08em
    T\kern-.1667em\lower.7ex\hbox{E}\kern-.125emX}}
\usepackage{balance}
\begin{document}
\title{VP-AutoTest: A Virtual-Physical Fusion Autonomous Driving \\Testing Platform}

\author{Yiming Cui, Shiyu Fang, Jiarui Zhang, Yan Huang, Chengkai Xu \\ Bing Zhu, Hao Zhang, Peng Hang,~\IEEEmembership{Senior Member,~IEEE,} and Jian Sun,~\IEEEmembership{Senior Member,~IEEE}

\thanks{This work was supported in part by the National Natural Science Foundation of China (52125208, 62433014), the Shanghai Scientific Innovation Foundation (No.23DZ1203400), and the Fundamental Research Funds for the Central Universities.}
\thanks{Y. Cui, S. Fang, J. Zhang, Y. Huang, C. Xu, P. Hang and J. Sun are with Department of Traffic Engineering \& Key Laboratory of Road and Traffic Engineering, Ministry of Education, Tongji University, Shanghai, China. (e-mail: \{yim211, fangshiyu, zjr0915, huangyan520, xuchengkai, hangpeng, sunjian\}@tongji.edu.cn).}
\thanks{H. Zhang is with the College of Electronic and Information Engineering, Department of Control Science and Engineering, and Shanghai Institute of Intelligent Science and Technology, Tongji University, Shanghai 200092, China. (e-mail: zhang\_hao@tongji.edu.cn).}
\thanks{B. Zhu is State Key Laboratory of Automotive Simulation and Control, Jilin University, Changchun 130022, China. (e-mail: zhubing@jlu.edu.cn).}
\thanks{Corresponding author: Jian Sun}
}


\maketitle

\begin{abstract}
The rapid development of autonomous vehicles has led to a surge in testing demand. Traditional testing methods, such as virtual simulation, closed-course, and public road testing, face several challenges, including unrealistic vehicle states, limited testing capabilities, and high costs. These issues have prompted increasing interest in virtual-physical fusion testing. However, despite its potential, virtual-physical fusion testing still faces challenges, such as limited element types, narrow testing scope, and fixed evaluation metrics. 
To address these challenges, we propose the Virtual-Physical Testing Platform for Autonomous Vehicles (VP-AutoTest), which integrates over ten types of virtual and physical elements, including vehicles, pedestrians, and roadside infrastructure, to replicate the diversity of real-world traffic participants. The platform also supports both single-vehicle interaction and multi-vehicle cooperation testing, employing adversarial testing and parallel deduction to accelerate fault detection and explore algorithmic limits, while OBU and Redis communication enable seamless vehicle-to-vehicle (V2V) and vehicle-to-infrastructure (V2I) cooperation across all levels of cooperative automation. Furthermore, VP-AutoTest incorporates a multidimensional evaluation framework and AI-driven expert systems to conduct comprehensive performance assessment and defect diagnosis. Finally, by comparing virtual–physical fusion test results with real-world experiments, the platform performs credibility self-evaluation to ensure both the fidelity and efficiency of autonomous driving testing. Please refer to the website for the full testing functionalities on the autonomous driving public service platform OnSite: \href{https://www.onsite.com.cn/}{https://www.onsite.com.cn/}.
\end{abstract}


\section{Introduction}
\label{section:I}
Autonomous Driving (AD) has entered the stage of commercial deployment \cite{shi2024soar}. However, various issues still arise in open-road operations \cite{di2024comparative}, prompting manufacturers to continuously update their algorithm versions through Over-The-Air (OTA) mechanisms \cite{ghosal2022secure}. This iterative process has led to a steadily increasing demand for AD testing. For academia, testing serves as a crucial means to evaluate the performance of AD algorithms and facilitate their application in real-world scenarios. From an industrial standpoint, testing remains an indispensable step prior to deployment and large-scale adoption. Nevertheless, due to the significant differences in testing requirements, most existing platforms are designed for specific functions and lack scalability. As a result, there is still no standardized or comprehensive testing framework for AD. Therefore, developing an integrated and efficient testing platform has become an urgent and critical challenge.

The International Organization of Motor Vehicle Manufacturers (OICA) identified virtual simulation, closed-track, and public road testing as the three pillars of validation \cite{unece2019future}. Broadly, these can be categorized into virtual-environment and real-world testing approaches. Virtual simulation platforms like PreScan \cite{shibo2025simulation}, CARLA \cite{Dosovitskiy17}, and Apollo \cite{fan2018baidu} construct traffic flow models to enable interaction with the tested algorithms, allowing for performance evaluation in a controlled environment. Although simulations offer low-cost, safe, and repeatable testing environments that support rapid iteration \cite{yan2021distributionally, sun2021corner}, they suffer from inherent discrepancies between simulated dynamics and real-world behaviors \cite{stocco2022mind}. In contrast, real-world testing provides authentic physical feedback \cite{koopman2016challenges}. Closed-course tests offer controlled conditions but are limited by fixed layouts and narrow functional coverage. Public road testing better reflects real traffic environments yet involves high costs, complex regulations, and safety risks. Consequently, relying solely on either virtual or real-world testing remains insufficient for comprehensive AD validation.

Due to the limitations of relying solely on virtual or real-world testing, researchers have explored hybrid testing methods that combine the convenience of simulations with the authenticity of physical verification. 
In recent years, concepts such as mixed reality (MR) \cite{szalai2020mixed, chen2020mixed, zofka2018traffic}, augmented reality (AR) \cite{feng2018augmented, weiguo2024augmented}, and digital twin (DT) \cite{2019DigitalDefination, schwarz2022role, duan2024digital, stocco2022mind, ai2023pmworld} have gradually been applied to the autonomous driving testing domain to construct virtual–physical fusion testing environments for typical traffic scenarios of research interest, such as merge areas \cite{yu2022autonomous}, roundabouts \cite{li2024digital, shahriar2024digital}, intersections \cite{samak2024metaverse, yu2025digital, dasgupta2021transportation}, and adverse weather conditions like rain and fog \cite{neto2025twice}, supporting the evaluation of autonomous driving functional performance. 
\citet{feng2018augmented} developed an AR environment by integrating virtual traffic flows from VISSIM with real connected vehicles, enabling the evaluation of cooperative driving behaviors under mixed real–virtual interactions.
Building upon this concept, \citet{weiguo2024augmented} proposed an AR-based platform that combined simulated and real sensor data to enhance real-world scenarios, focusing on the testing of perception and decision-making algorithms.
\citet{campolo2024edge} introduced a digital twin framework that synchronized real vehicles with their virtual counterparts through edge-cloud architecture, enabling real-time, bidirectional interaction for high-fidelity testing.
However, despite the potential of virtual-physical fusion approaches, current research and applications still face several limitations: \textbf{1) Limited element variety}: The number and types of elements involved in existing virtual-physical fusion tests are relatively few. \textbf{2) Narrow testing scope}: Testing functions are often focused on specific algorithm validation or system evaluation, lacking comprehensiveness. \textbf{3) Simplistic evaluation methods}: Existing evaluation methods are often rigid, relying on specific performance metrics without considering deeper interactivity or multidimensional assessments.
Therefore, while virtual-physical fusion testing plays a significant role in improving testing efficiency and safety, there is a need to expand the functionality and diversity of testing and to explore more flexible and comprehensive evaluation standards to meet the growing complexities of AD validation.

To address the aforementioned challenges, we propose the Virtual-Physical Testing platform for Autonomous vehicles (VP-AutoTest). By integrating over ten distinct virtual-physical elements, it enables both single-vehicle and multi-vehicle testing functionalities. Additionally, it incorporates multidimensional evaluation and defect diagnosis, thus supporting comprehensive and systematic testing of AD systems. The key contributions of the proposed platform are summarized as follows.

\begin{enumerate}
\item \textbf{Integration of Diverse Virtual-Physical Elements}: To replicate the complexity and diversity of real-world traffic participants, the proposed VP-AutoTest integrates a wide range of physical and virtual components. The physical elements include Connected and Automated Vehicles (CAVs), cloud-controlled vehicles, mannequins, and roadside infrastructure. The virtual components consist of simulated CAVs, remotely operated vehicles controlled by driving simulators, and simulated background traffic flows, enabling a comprehensive and realistic testing environment.

\item \textbf{Single and Multi-Vehicle Testing Capabilities}: To efficiently evaluate the various capabilities of AD systems, platform supports both single-vehicle interaction and multi-vehicle cooperation tests. For single-vehicle interaction, we introduce adversarial testing to adjust challenge intensity based on system performance, and parallel deduction to simulate high-risk takeover scenarios for evaluating decision-making algorithms. In the multi-vehicle cooperation tests, the platform utilizes communication between OBUs and Redis to evaluate V2V and V2I cooperation, covering the entire spectrum of Cooperative Driving Automation (CDA).

\item \textbf{Multidimensional Evaluation and Defect Diagnosis}: The platform facilitates multidimensional comparisons of algorithms, both horizontally and vertically, enabling detailed performance assessments across various versions. Additionally, it incorporates AI-driven expert systems that leverage knowledge reasoning to generate comprehensive diagnostic reports for the Algorithm Under Test (AUT), providing constructive insights for algorithmic improvement and optimization.

\item \textbf{Platform Credibility Self-Evaluation}: Since credibility remains one of the most critical challenges in virtual–physical fusion testing, VP-AutoTest incorporates a credibility self-evaluation mechanism that quantitatively assesses the reliability of each test case. Based on credibility analysis, the platform adaptively adjusts element combinations—adding more physical components in low-credibility scenarios to enhance realism and increasing virtual components in high-credibility ones to improve efficiency, thereby enhancing testing efficiency while ensuring credibility.

\end{enumerate}

The remainder of this paper is organized as follows. Section \ref{section:II} briefly introduces the overall architecture of the platform and the key technologies employed in its construction. Section \ref{section:III} describes the testing capabilities of the platform and relevant case studies in detail. Section \ref{section:IV} introduces the platform’s evaluation capabilities, including both the performance of the tested algorithms and the reliability of the platform itself. Section \ref{section:V} concludes the paper and outlines potential directions for future enhancement of the platform.

\section{Overview of the Proposed VP-AutoTest Framework} %
\label{section:II}

\begin{figure*}[htbp]
  \begin{center}
  \centerline{\includegraphics[width=7in]{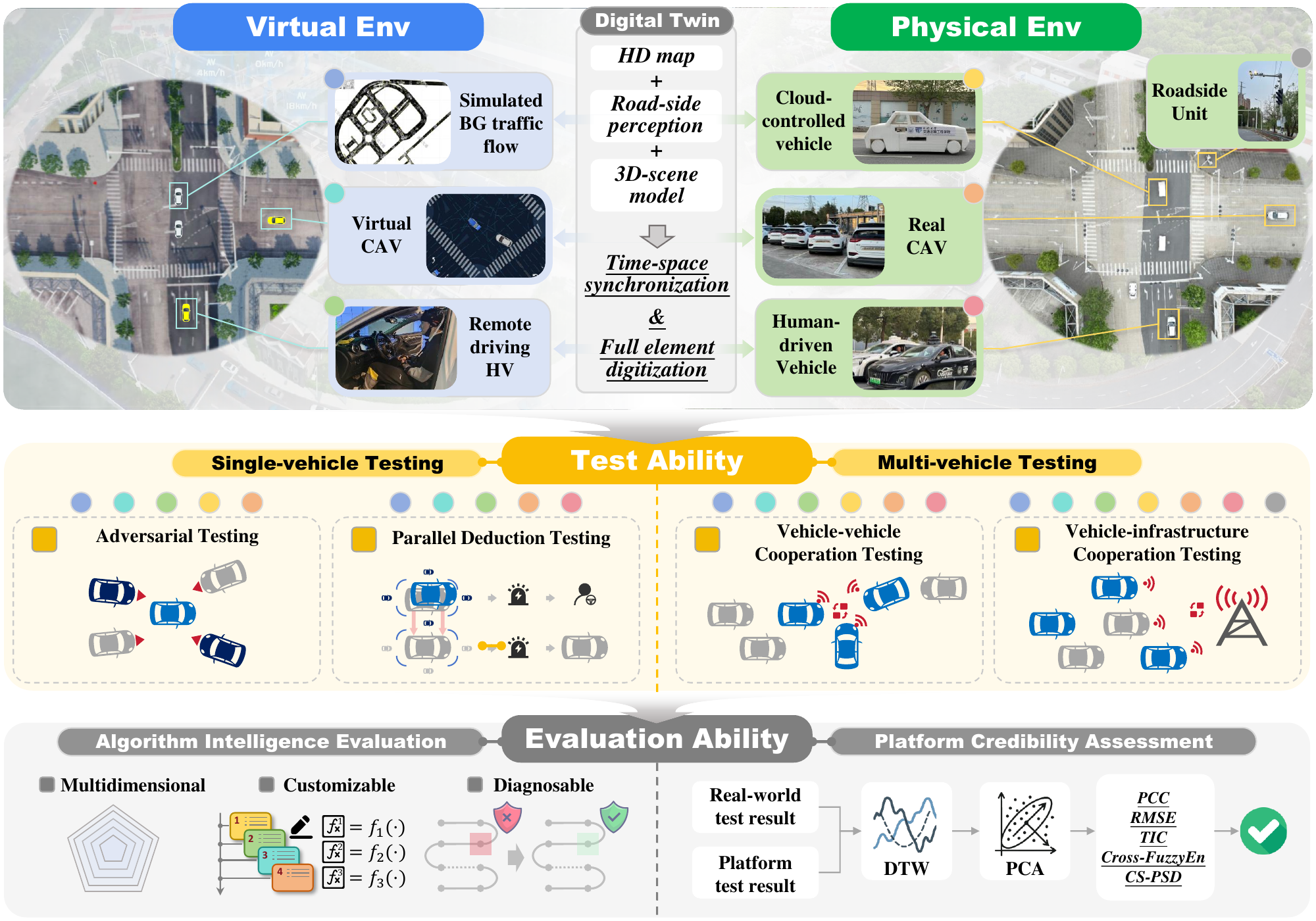}}
  \caption{Framework of the virtual-physical fusion testing platform VP-AutoTest.}\label{Framework}
  \end{center}
  \vspace{-0.4cm}
\end{figure*}

The proposed VP-AutoTest is designed to achieve seamless integration between the physical and virtual domains, as illustrated in Fig.~\ref{Framework}. Built upon digital twin technology, the platform establishes a spatiotemporally aligned digital twin base that provides a unified reference for virtual–physical fusion testing. Static elements are digitally reconstructed based on high-definition maps to form the spatial reference framework of the test site. Dynamic elements are captured and localized in real time through perception and sensing devices deployed across the test field. Simulated elements are synchronized with real-world perception data, enabling fusion at the perception layer. Together, these three types of elements ensure high-fidelity synchronization between the real and virtual worlds, thereby supporting unified multi-source data processing, accurate entity reconstruction, and reliable algorithm validation.

Specifically, in terms of temporal synchronization, VP-AutoTest supports multiple clock alignment standards according to application requirements. The Network Time Protocol (NTP) achieves a synchronization accuracy of 10~ms, while the Precision Time Protocol (PTP) provides accuracy up to 50~ns, and Global Navigation Satellite System (GNSS) synchronization reaches 10~ns. For spatial synchronization, a centimeter-level high-definition map is used to construct a unified spatial reference system. The three-dimensional scene reconstruction service integrates physical features and virtual models and is regularly updated to ensure consistency between domains.

To optimize the allocation of virtual and physical elements and improve resource utilization, VP-AutoTest introduces a human–vehicle–road–environment coupling risk-field model to quantify the influence of background traffic on the test target. A risk contribution evaluation system is then established to determine the optimal distribution of virtual and physical elements based on risk gradient analysis, ensuring economical resource deployment and maximal risk coverage. Combined with the platform’s credibility self-evaluation mechanism, the virtual–physical composition is dynamically adjusted across iterative tests, maintaining testing credibility while enhancing efficiency.

As shown in Fig.~\ref{Framework}, the physical test field includes a variety of real-world driving scenarios such as parking lots, straight roads, intersections, and roundabouts. The virtual environment, built upon the high-definition map (HD map), achieves a one-to-one digital replica of these physical scenes. Through the fusion of virtual environments that generate controllable background traffic flows, virtual CAVs, and remotely operated HDVs, the platform enables free combinations and dynamic interactions among virtual and physical participants. This design supports both single-vehicle and multi-vehicle testing requirements. The single-vehicle testing includes adversarial testing and parallel deduction, which focus on evaluating decision-making performance and system robustness. The multi-vehicle testing involves V2V and V2I cooperation tests to assess cooperative driving capabilities under complex traffic conditions.
In addition, the platform provides comprehensive evaluation capabilities for both algorithm performance and its own credibility. It supports multidimensional and customizable assessments of algorithmic intelligence, enabling defect diagnosis and targeted improvement. Meanwhile, the platform can also self-evaluate its credibility to ensure its reliability and usability.

\section{Virtual \& Physical Components}
\label{section:III}
\begin{figure}[htbp]
  \begin{center}
  \centerline{\includegraphics[width=3.5in]{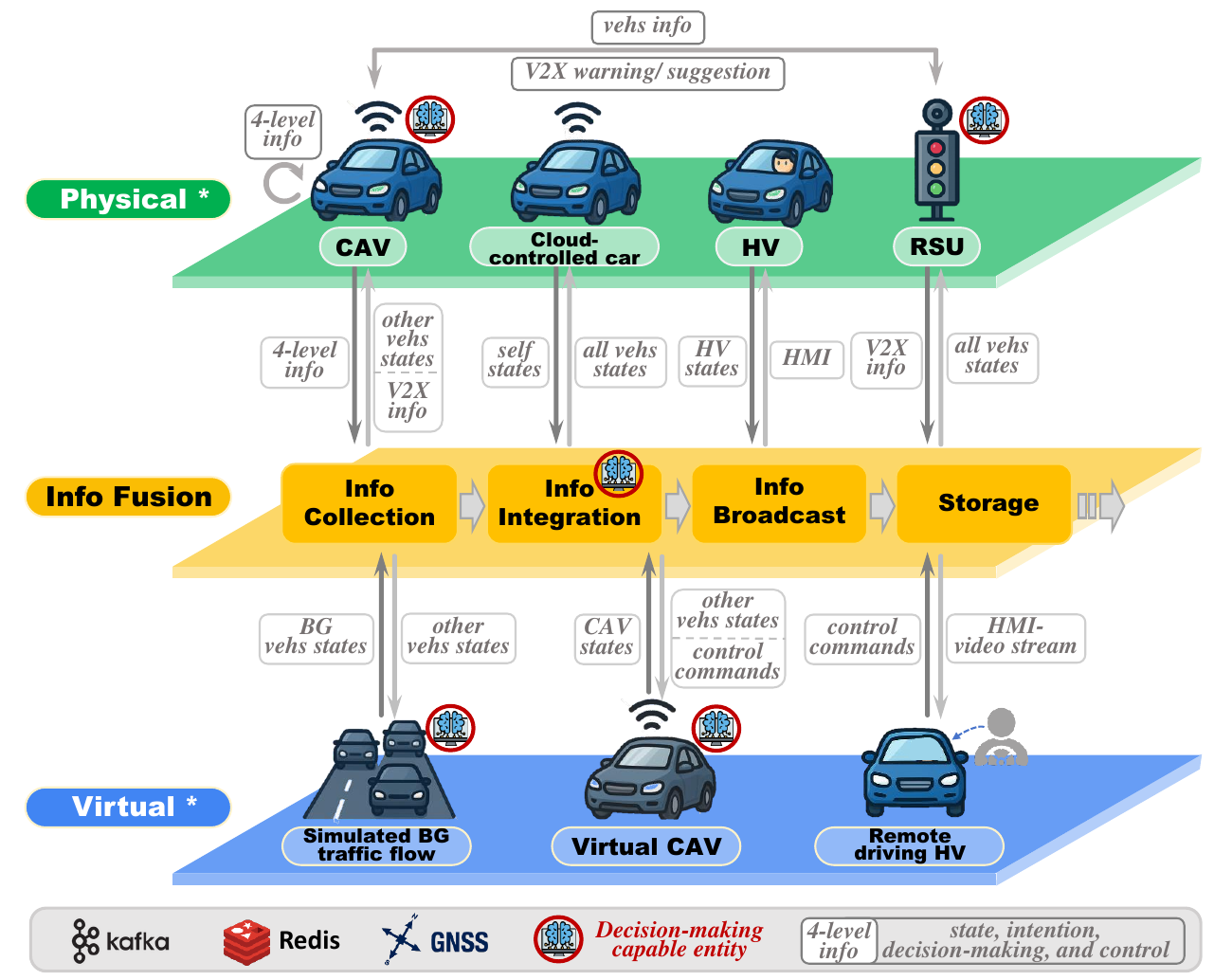}}
  \caption{Data transmission pipeline of the VP-AutoTest.}\label{Workflow}
  \end{center}
  \vspace{-0.4cm}
\end{figure}

The testing elements are categorized into two types, physical testing elements and virtual testing elements, based on whether the main entities have a physical presence in the physical testing site.
This section provides a detailed description of each virtual and physical component, including their composition, operating principles, communication mechanisms, and other relevant technical details. The data transmission pipeline for components of the platform is shown in Fig.~\ref{Workflow}.

\subsection{Physical Components}
\subsubsection{CAV}
\begin{figure}[htbp]
  \begin{center}
  \centerline{\includegraphics[width=3.5in]{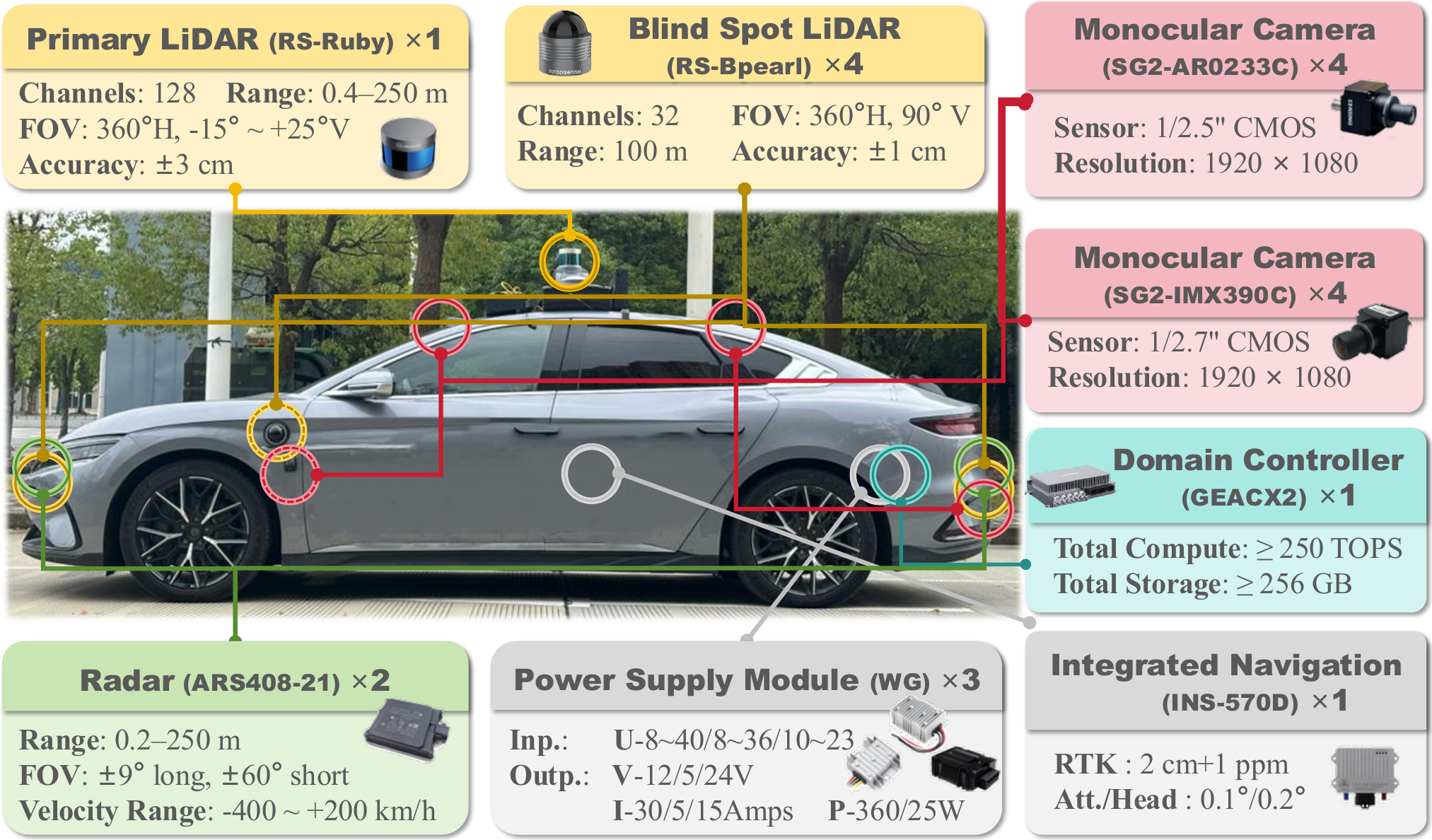}}
  \caption{Overview of the connected and automated vehicle setup.}\label{CAV}
  \end{center}
  \vspace{-0.4cm}
\end{figure}

In VP-AutoTest, CAVs typically serve as the primary test agents for various test tasks. A total of 14 CAVs have been integrated into the platform, supporting algorithm evaluation across multiple levels of automation. The fleet includes models such as Hongqi ES3, Hongqi EQM5, BYD Han, and Aion LX80. Each vehicle is equipped with a full-stack sensor suite—one primary LiDAR, four blind-spot LiDAR units, two millimeter-wave radars, and eight monocular cameras. The detailed hardware configuration is shown in Fig.~\ref{CAV}.

For platform-side communication, CAVs interface with the system through Redis \cite{carlson2013redis}, enabling real-time access to task information, test-start commands, and the positions of virtual vehicles in the environment. For interactions with other physical entities, vehicle-to-everything (V2X) communication is achieved via OBU broadcasting, through which CAVs can dynamically obtain the positions of other cooperative vehicles or the status and perception data from roadside intelligent units during multi-vehicle tests.
%

\subsubsection{Cloud-controlled vehicle}
The cloud-controlled vehicle consists of a real drive-by-wire chassis, a domain controller, and a foam exterior. Its deformable shell and driverless operation make it particularly suitable for high-risk test scenarios. It supports remote driving, trajectory tracking, and active collision avoidance. The drive-by-wire chassis enables both direct chassis control and real-time access to vehicle dynamics data, while the domain controller serves as the onboard computing unit. In addition, positioning is provided by an integrated GNSS navigation system, and a 5G-CPE module ensures communication between the domain controller and the VP-AutoTest. Through Redis-based real-time communication with the platform, cloud-controlled vehicles can participate in adversarial, cooperative, and other multi-agent test tasks.

\subsubsection{Human-driven vehicle (HDV)}
In VP-AutoTest, HDVs involved in testing fall into two categories. The first category consists of CAVs switched to manual driving mode. Although the vehicles are manually operated, they can still communicate with the platform through Redis and OBU modules, with their status labeled as manual. The second category includes unmodified HDVs without any additional sensing or communication hardware. Because these vehicles lack communication modules, they cannot directly exchange information with the platform or other vehicles. Instead, they are dynamically detected and tracked by intelligent roadside units deployed in the test field, and their digital twins are reconstructed in the virtual environment. Therefore, other vehicles can then obtain the states of these HDVs through Redis, with the real-time discrepancy between physical perception and digital twin representation kept within 10 cm.

\subsubsection{Road side unit (RSU)}
The RSUs deployed in VP-AutoTest include video detectors, millimeter-wave radars, roadside LiDARs, edge computing units, variable lane controllers, variable lane indicators, and variable signal heads. They provide a communication latency of less than 200 ms and a coverage range exceeding 1km, while also supporting tunnel-mode networking and high-precision positioning through 5G communication. The RSUs broadcast information on surrounding obstacles and can retrieve Signal Phase and Timing (SPAT) messages from signal controllers. These messages are then delivered to the vehicles, enabling timely hazard warnings and advisory strategies that enhance driving safety and efficiency.

\subsection{Virtual Components}
\subsubsection{Virtual CAV}
In the proposed VP-AutoTest, domain controllers can function as virtual CAVs to perform testing tasks, enabling efficient pre-evaluation of algorithm performance and serving as an essential prerequisite before real-vehicle testing. Specifically, the platform transmits surrounding-vehicle motion information from the virtual–physical fusion environment to the domain controller, which then makes real-time driving decisions based on the received data, updates the state of the virtual CAV, and sends the results back to the platform to support interactive testing. Through standardized communication protocols, VP-AutoTest supports the integration of domain controllers from multiple major vendors, including NVIDIA, China Automotive Innovation Corporation (CAIC), and Huawei.

\subsubsection{Remote driving}
In addition, drivers can participate in testing through remote driving, providing a safer means to control real or virtual vehicles while interacting with the AUT. Specifically, driving simulators are integrated into the proposed VP-AutoTest to receive state information from surrounding environment. Based on the driver’s real-time inputs, the system can control either a physical vehicle in the test field or a virtual vehicle in the digital twin environment.

\subsubsection{Background traffic flow}
Virtual background traffic flow is generated using the microscopic traffic simulation software TESSNG \cite{zhao2023development}, capable of accommodating testing scenarios involving diverse traffic participants, extended durations, and high traffic volumes. TESSNG provides comprehensive secondary development functionality, enabling customized modifications to traffic participant attributes, core traffic simulation models, and generation logic through its APIs. The generated virtual background traffic is uploaded to the platform via Redis, while the platform concurrently transmits data from other components back to TESSNG. This data is mapped to form virtual vehicles within the background traffic flow, enabling real-time interaction between the mapped virtual vehicles and the simulated background traffic.

\section{Testing Ability}
\label{section:IV}
This chapter systematically presents the testing capabilities supported by the virtual-physical fusion testing platform. These capabilities are categorized into single-vehicle and multi-vehicle virtual-physical fusion testing, with each category elaborating on the corresponding testing objectives, technical implementation methodologies, and representative test cases conducted based on the platform. The videos of cases are displayed here \footnote{https://yimcui.github.io/VP-AutoTest/}.

\subsection{Single-vehicle Virtual-physical Fusion Testing}

Testing the individual capabilities of AVs has long been a central task for manufacturers and a prerequisite for safe deployment \cite{lou2022testing}. Although current algorithms can be well validated in routine scenarios, their performance boundaries in edge cases remain insufficiently examined due to the extreme rarity of such events in naturalistic data—precisely where AVs are most likely to fail \cite{jung2022automatic}. This gap highlights the need for targeted evaluation of AV behavior in low-frequency, high-risk situations.

To this end, the proposed VP-AutoTest platform offers two key capabilities for single-vehicle testing. 1) Adversarial testing: adaptively generating adversarial behaviors for surrounding vehicles to rapidly expose the capability boundaries of the AUT. 2) Parallel deduction testing: when a takeover occurs, running the AUT in parallel to assess whether it could in fact handle the current scenario, thereby identifying and correcting false-positive takeovers.

\subsubsection{Adversarial Testing}



\begin{figure}[htbp]
  \begin{center}
  \centerline{\includegraphics[width=3.5in]{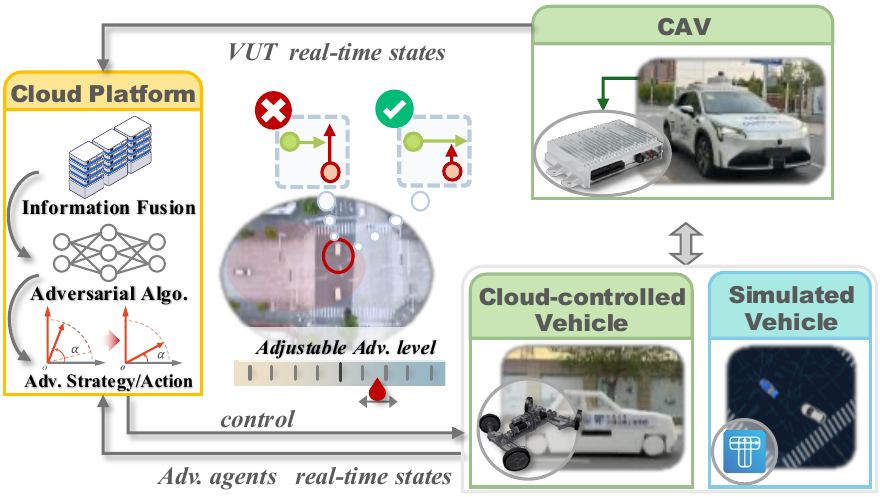}}
  \caption{The implementation schematic diagram of adversarial testing.}\label{AT-SD}
  \end{center}
  \vspace{-0.4cm}
\end{figure}

Adversarial testing is a method that evaluates the performance of the VUT in extreme scenarios by adaptively adjusting adversarial strategies according to the VUT’s real-time behavior. Its primary goal is to efficiently and systematically expose critical weaknesses that may be difficult to uncover through standard testing procedures, thereby enabling a more effective exploration of the capability boundaries of the algorithm under test \cite{guo2025interactive}. As illustrated in Fig.~\ref{AT-SD}, the state information of the VUT, including its precise position, velocity, and acceleration, is continuously transmitted to the adversarial testing algorithm deployed in the cloud. The algorithm then incorporates the current state of surrounding background vehicles and generates adversarial behaviors for a selected subset of these vehicles using methods such as reinforcement learning or game-theoretic optimization, including aggressive overtaking, lane straddling, continuous lane changes, and sudden emergency braking, among others. These background vehicles may consist of virtual simulation entities or physical vehicles operating in the test field.

In addition, the adversarial algorithm adaptively adjusts the strength of adversarial behaviors based on the VUT’s real-time responses. For example, when the VUT demonstrates strong safety performance, the algorithm increases the aggressiveness of background vehicles in order to create more safety-critical scenarios and further evaluate the limits of the autonomous driving system. 

\begin{figure}[htbp]
  \begin{center}
  \centerline{\includegraphics[width=3.5in]{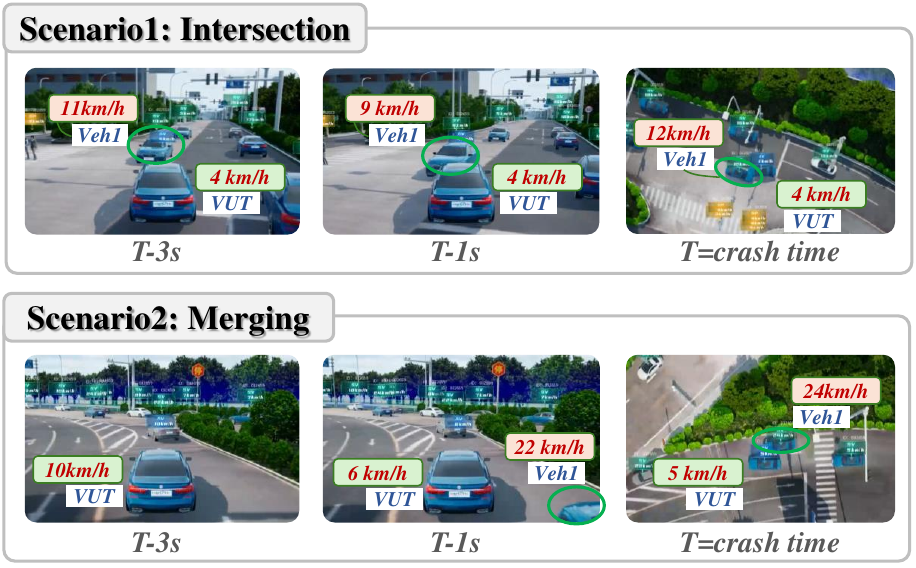}}
  \caption{Typical testing cases of adversarial testing.}\label{AT-case}
  \end{center}
\end{figure}

Fig.~\ref{AT-case} illustrates two representative adversarial test cases generated by the adversarial method proposed in \cite{mei2025llm} while AUT adopts the reservation-based method rotational projection IDM approach  \cite{xu2018distributed}.
In the scenario-1, Veh1 exhibits a rushing behavior during a conflict at an intersection. During the VUT’s left-turn maneuver, the PIDM tends to prioritize the VUT’s passage because it is closer to the conflict point. However, the straight Veh1 increases its adversarial intensity by accelerating from 9 km/h to 12 km/h in an attempt to seize the right of way. The AV does not decelerate to yield as expected, indicating that the algorithm struggles with this complex interaction.
In the scenario-2, Veh1 serves as the interacting agent on the target lane during the VUT’s merging process. As the AV attempts to merge, Veh1 displays an aggressive acceleration pattern. The VUT fails to leave sufficient safety space for this vehicle, ultimately resulting in a sideswipe collision.

\begin{figure}[htbp]
  \begin{center}
  \centerline{\includegraphics[width=3.3in]{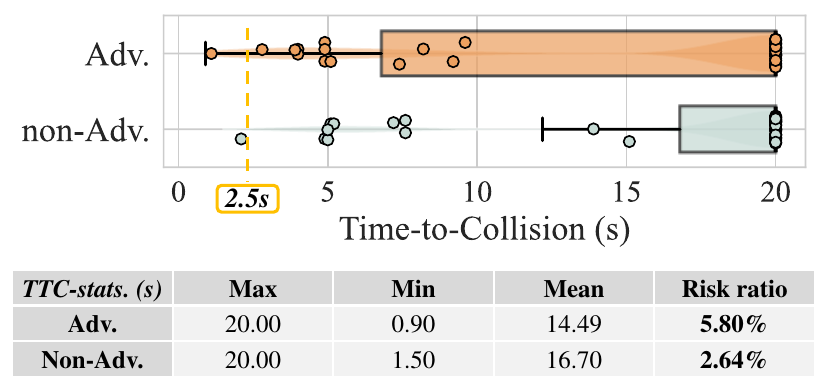}}
  \caption{TTC variation curves of VUT under adversarial and non-adversarial tests.}\label{AT-TTC}
  \end{center}
\end{figure}

Furthermore, Fig.~\ref{AT-TTC} illustrates the changes in the two-dimensional Time-to-Collision (TTC) distribution before and after the adaptive adjustment of the adversarial strategy during testing, where lower values indicate higher collision risk. The results show that, under adversarial testing, the frequency of hazardous events (defined as \(TTC < 2.5 s\)) increases significantly, from 2.64\% to 5.80\%. This confirms that the adversarial testing capability of the virtual–physical fusion platform can effectively create more challenging and hazardous test conditions, thereby substantially improving the efficiency of safety validation.

\subsubsection{Parallel Deduction Testing}
During real-world AD tests, a safety driver is typically required to take control whenever the vehicle shows signs of unsafe or abnormal behavior. Although such interventions ensure testing safety, limited trust in the AUT often leads to excessive or unnecessary takeovers, preventing a full assessment of the algorithm’s true capability. 
During real-world AD tests, a safety driver is typically required to take control whenever the vehicle exhibits signs of hazardous or improper behavior. While such interventions ensure operational safety, insufficient trust in the AUT often leads to excessive or unnecessary takeovers. Statistics indicate that 77\% of takeovers are proactively initiated by the driver rather than being triggered by system failures \cite{kohanpour2025trends, dixit2016autonomous}. This prevalence of discretionary interventions prevents a comprehensive assessment of the AUT's true capabilities.

\begin{figure}[htbp]
  \begin{center}
  \centerline{\includegraphics[width=3.5in]{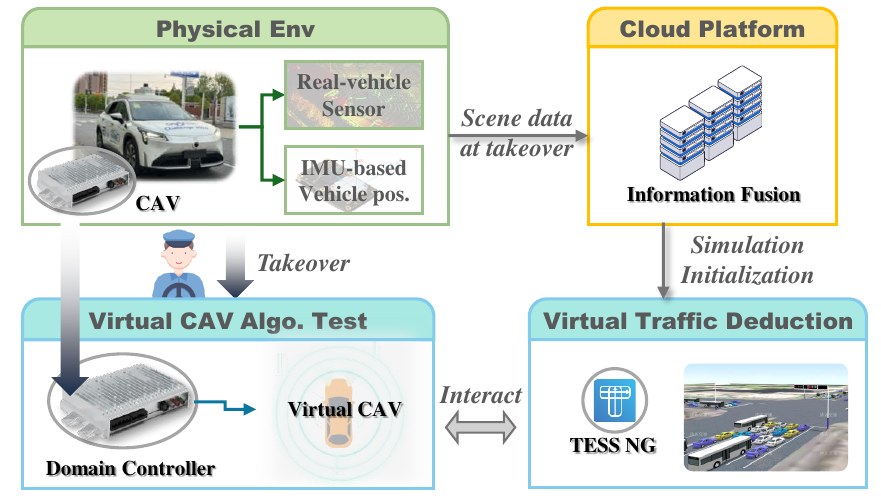}}
  \caption{The implementation schematic diagram of parallel deduction testing.}\label{PD-SD}
  \end{center}
  \vspace{-0.4cm}
\end{figure}

To address this issue while still avoiding real-world risks, the platform introduces a parallel deduction method that evaluates whether the AUT could have successfully handled the scenario without human intervention, as shown in Fig.~\ref{PD-SD}. Specifically, once the VUT is taken over and switched to manual mode, the controller sends the scene data at the takeover moment to TESSNG. After receiving the data, TESSNG continuously simulates the subsequent background traffic flow, enabling the AUT to interact with the evolving virtual environment within the virtual world. This process allows the system to determine whether the algorithm is capable of correctly handling the scenario without requiring an actual on-road attempt.

Fig.~\ref{PD-SD} presents a parallel deduction test case in which a reinforcement learning–based algorithm \cite{xu2025tell} performs an unprotected left turn at a T-junction. At the beginning, the VUT operates normally, but as the test progresses, both the VUT and Veh1 move closer to the conflict point. Considering the rising level of risk, the safety operator takes over and performs a deceleration-yield action, allowing the opposing straight vehicle to pass first at a speed of 19 km/h. From the moment of takeover, the platform continues the closed loop simulation in the digital twin environment using the onboard AD algorithm. The simulation shows that the algorithm could have completed a priority passing maneuver in this scenario, achieving efficient traversal with a smaller deceleration, while the opposing vehicle would have slowed down to yield. This demonstrates the algorithm’s capability in scenarios involving inappropriate human takeover.

\begin{figure}[htbp]
  \begin{center}
  \centerline{\includegraphics[width=3.5in]{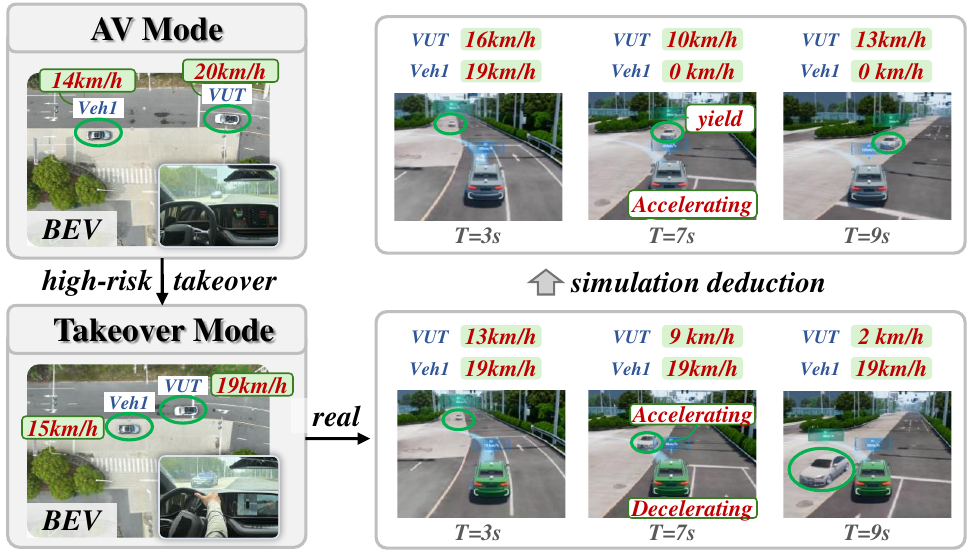}}
  \caption{Parallel deduction testing case.}\label{case_2}
  \end{center}
  \vspace{-0.4cm}
\end{figure}

\subsection{Multi-vehicle Virtual-physical Fusion Testing}


During the trial operation phase of autonomous driving, single-vehicle intelligence algorithms often face safety and efficiency issues in complex conflict scenarios due to limited decision-making capabilities. With the advancement of vehicular communication technologies, V2V and V2I cooperation has become feasible, offering a promising pathway to enhance AV performance in complex interactive environments \cite{jia2016enhanced}. The SAE J3216 standard further categorizes CDA into four levels, providing technical specifications for research in cooperative driving \cite{fang2025towards}.

To support cooperative capability testing, the platform leverages the proposed spatiotemporal synchronization technology to ensure millisecond-level communication among participating agents. Based on the type of cooperation involved, the tests can be categorized into V2V and V2I scenarios.

\subsubsection{Vehicle-vehicle Cooperation Testing}



Vehicle-to-vehicle cooperation has the potential to significantly enhance CAV performance in complex scenarios. However, existing research on cooperative driving remains largely confined to simulation, primarily due to stringent communication stability requirements and the limited availability of cooperative hardware resources. To bridge the gap between simulation-based demonstrations and real-world deployment, the proposed VP-AutoTest aligns virtual and physical vehicles within a unified space through spatiotemporal synchronization, enabling scalable multi-vehicle interaction and cooperation. Communication among physical vehicles is handled through OBUs, while communication between virtual and physical elements is achieved via Redis.

Following the SAE J3216 standard and the types of information exchanged in V2V communication, the vehicle-to-vehicle cooperation testing module evaluates four CDA levels: state sharing, intent sharing, cooperative decision-making, and cooperative control. When vehicles exchange only basic states like positions, the system assesses communication stability at the state-sharing level. When desired decisions are also shared, it evaluates whether consensus algorithms can reach agreement under intent sharing. When vehicles convert these inputs into semantic decisions, cooperative decision-making algorithms are examined for safety and efficiency. Finally, when control commands are exchanged, the robustness of cooperative control can be tested. Through these capabilities, VP-AutoTest enables comprehensive evaluation of cooperative driving algorithms across different scenarios and CDA levels.

\begin{figure}[htbp]
  \begin{center}
  \centerline{\includegraphics[width=3.5in]{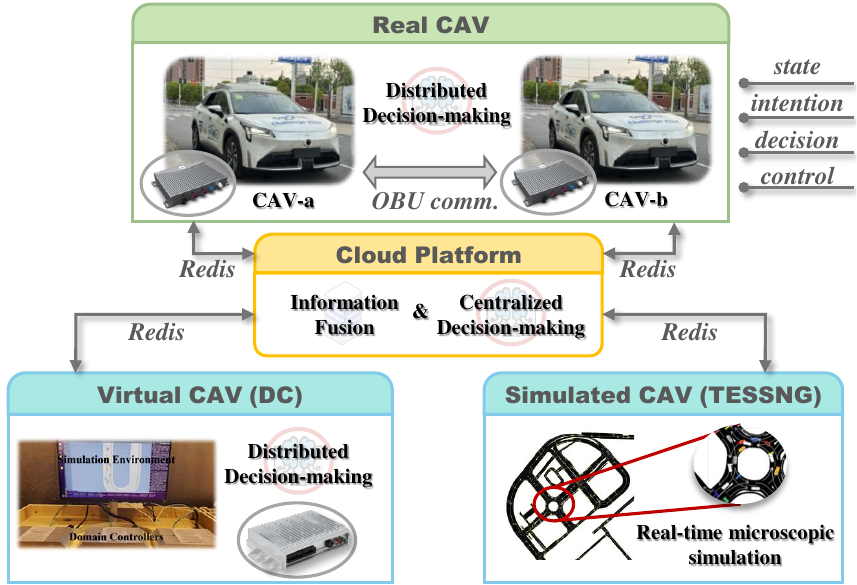}}
  \caption{The implementation schematic diagram of vehicle-vehicle cooperation.}\label{v2v-SD}
  \end{center}
\end{figure}

\begin{figure}[htbp]
  \begin{center}
  \centerline{\includegraphics[width=3.5in]{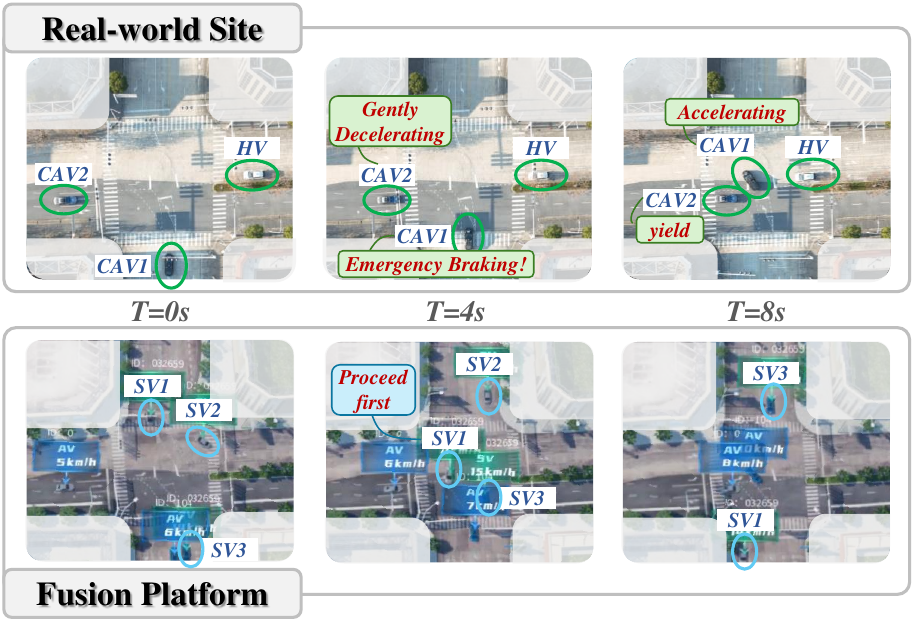}}
  \caption{Vehicle-vehicle cooperation with cooperative decision-making testing case.}\label{v2v-decision}
  \end{center}
  \vspace{-0.4cm}
\end{figure}

Fig.~\ref{v2v-decision} presents a case study with three physical and three virtual vehicles, further validating the cooperative decision-making method in \cite{fang2023potential}. Although all vehicles ultimately passed the intersection safely, an unexpected braking occurred around 4 seconds when CAV1 suddenly stopped mid-intersection. A closer comparison between simulation and real execution shows that real vehicles place greater emphasis on comfort: CAV2 decelerated more gently than in simulation, unintentionally triggering CAV1’s emergency response. This finding highlights that, despite strong simulation performance, the method in \cite{fang2023potential} still requires refinement for deployment on real vehicles.

\subsubsection{Vehicle-infrastructure Cooperation Testing}

In addition, V2I cooperation has become a major research focus due to its potential to improve road safety, traffic efficiency, and management. However, despite rapid progress, challenges in fusing and aligning heterogeneous data mean that most V2I algorithms are still validated only in limited-scale settings. To address this, VP-AutoTest enables flexible configuration and integration of virtual and physical elements, including real vehicles, RSUs, MEC edge nodes, cloud-control platforms, and virtual background traffic, into a unified test environment. This design supports bidirectional control between vehicles and infrastructure across various V2I functions. According to test objectives, V2I tasks can be grouped into safety-oriented and efficiency-oriented categories: the former includes construction warnings, vulnerable road user alerts, and non-line-of-sight hazard detection; the latter includes congestion guidance and green-wave speed advisories.

In practice, MEC nodes perceive the motion of background vehicles, nearby CAVs, and pedestrians, and determine potential safety risks or inefficient behaviors relative to the ego vehicle. They then generate corresponding warnings or guidance messages and broadcast them via the OBU. Upon receiving these messages, the CAV displays them through the HMI and reacts according to its internal control logic, improving both safety and operational efficiency.

\begin{figure}[htbp]
  \begin{center}
  \centerline{\includegraphics[width=3.5in]{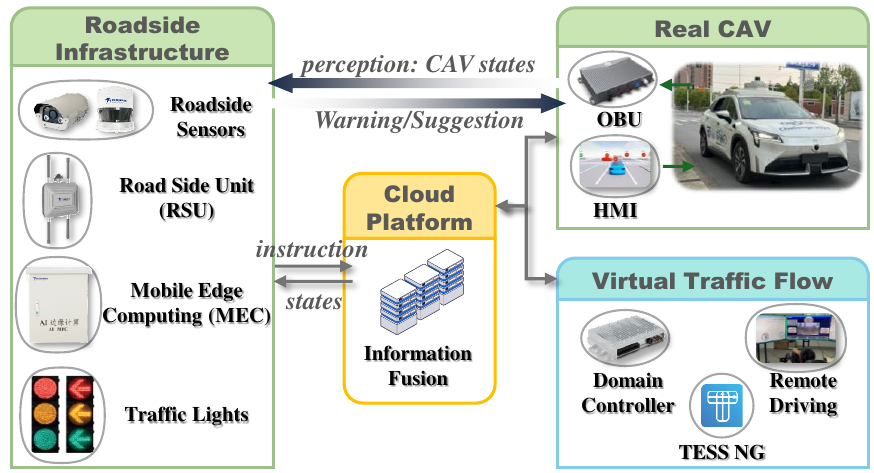}}
  \caption{The implementation schematic diagram of vehicle-infrastructure cooperation.}\label{v2i-SD}
  \end{center}
  \vspace{-0.4cm}
\end{figure}

\begin{figure}[htbp]
  \begin{center}
  \centerline{\includegraphics[width=3.5in]{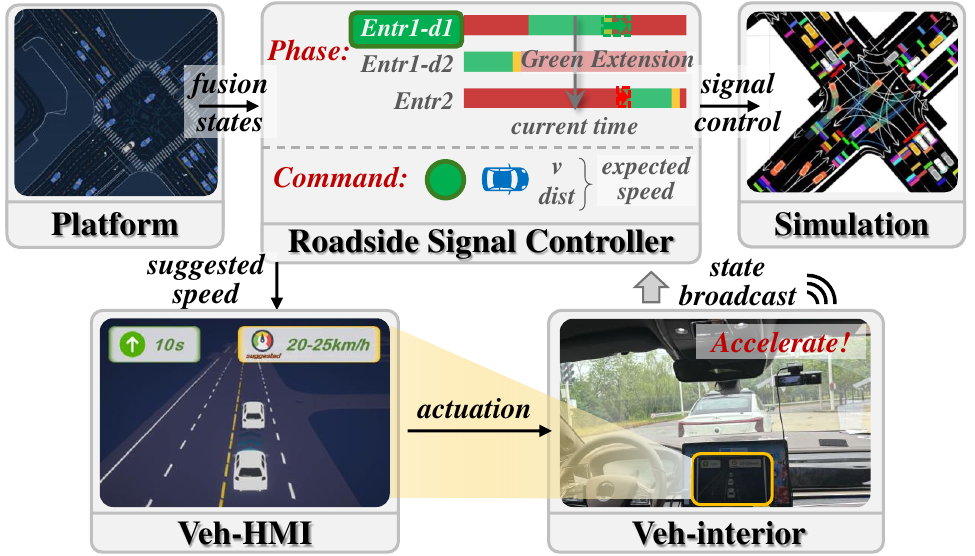}}
  \caption{Vehicle-signal cooperative control testing cases.}\label{case_4}
  \end{center}
  \vspace{-0.4cm}
\end{figure}

Fig.~\ref{case_4} illustrates a V2I-based cooperative control test scenario at a four-way intersection involving CAVs and traffic signal phases. Vehicles traveling along left-turn, straight, and right-turn paths are set up in all directions. The traffic signal system dynamically optimizes phase and sequence based on real-time traffic demand obtained from the platform, which includes both real and virtual vehicles \cite{li2021optimal}. For each active signal phase, V2I information is used to jointly optimize the CAV speed profiles and the duration of green lights. Finally, the roadside equipment issues recommended actions to vehicles, influencing their behavior to improve overall traffic efficiency, reduce queue lengths, and minimize vehicle delays.

\section{Evaluation Ability}
\label{section:V}
A systematic and comprehensive evaluation is a key step for identifying algorithmic weaknesses and potential improvement directions. The proposed VP-AutoTest conducts evaluation along two main dimensions: algorithm intelligence evaluation and platform credibility assessment.

\subsection{Algorithm Intelligence Evaluation}
Unlike conventional evaluations that focus on a single autonomous driving function, the platform establishes a multidimensional, integrated evaluation paradigm to comprehensively assess the intelligence level of autonomous driving systems. In addition, to support diverse testing and evaluation needs, the platform allows customization of evaluation objectives, schemes, and metrics. Finally, AI-driven expert systems analyze and diagnose vehicle performance during testing, providing actionable guidance for iterative improvement of both vehicles and algorithms.


\subsubsection{Multidimensional Comprehensive Evaluation}
Conventional evaluation methods often focus on single aspects of AD algorithms, which may lead to inadequate assessment in complex scenarios that require the coordinated involvement of multiple modules. To address this, the platform treats each vehicle as an integrated intelligent agent and performs a comprehensive intelligence assessment. The evaluation not only covers traditional dimensions such as safety, efficiency, and comfort, but also considers the vehicle’s potential impact on the surrounding environment, adding two additional dimensions—traffic law compliance and traffic coordination—to assess algorithmic intelligence. Traffic law compliance ensures strict adherence to regulations, while traffic coordination reflects the vehicle’s ability to interact and cooperate with other traffic participants. The multidimensional framework thus provides a more realistic assessment of overall vehicle performance. Moreover, the platform supports horizontal comparison of different algorithms in the same scenario and vertical comparison of the same algorithm across scenarios, facilitating adaptability analysis and optimal algorithm selection.


\subsubsection{Customizable Evaluation Requirements}
During AD testing, different scenarios and application requirements can significantly influence which aspects of vehicle performance are emphasized. To overcome the limitations of general evaluation frameworks in capturing scenario-specific characteristics, the platform allows customization of evaluation metrics. Testers can adjust or replace metrics according to actual needs. For example, in standard car-following or straight-driving scenarios, TTC effectively indicates vehicle safety, whereas in conflict-heavy scenarios such as roundabouts or intersections, Post-Encroachment Time (PET) may provide a more accurate assessment.
This approach enables the evaluation system to adapt dynamically to specific testing scenarios, producing intelligence assessments that better reflect actual testing demands rather than relying on fixed dimensions. Furthermore, users or regulatory bodies can configure the metric system based on their priorities, resulting in evaluation outcomes with greater interpretability and practical value.

\subsubsection{Evaluation Result Diagnosis and Analysis}
In the evaluation process, feedback plays a crucial role in guiding the continuous optimization and performance improvement of AD algorithms. Accordingly, the platform not only provides quantitative results across multiple dimensions but also supports algorithm diagnostics and improvement suggestions based on these results. To enhance the intelligence and specificity of the diagnostics, an AI expert module is incorporated, combining traditional expert rules with data-driven analysis. Through knowledge reasoning, it enables intelligent identification of algorithmic deficiencies and generates comprehensive diagnostic reports. By directly translating evaluation and diagnostic outcomes into actionable optimization insights, the platform establishes a tight closed loop between testing and algorithm iteration, thereby accelerating the performance enhancement of autonomous driving systems.

\begin{figure*}[htbp]
  \begin{center}
  \centerline{\includegraphics[width=6.8in]{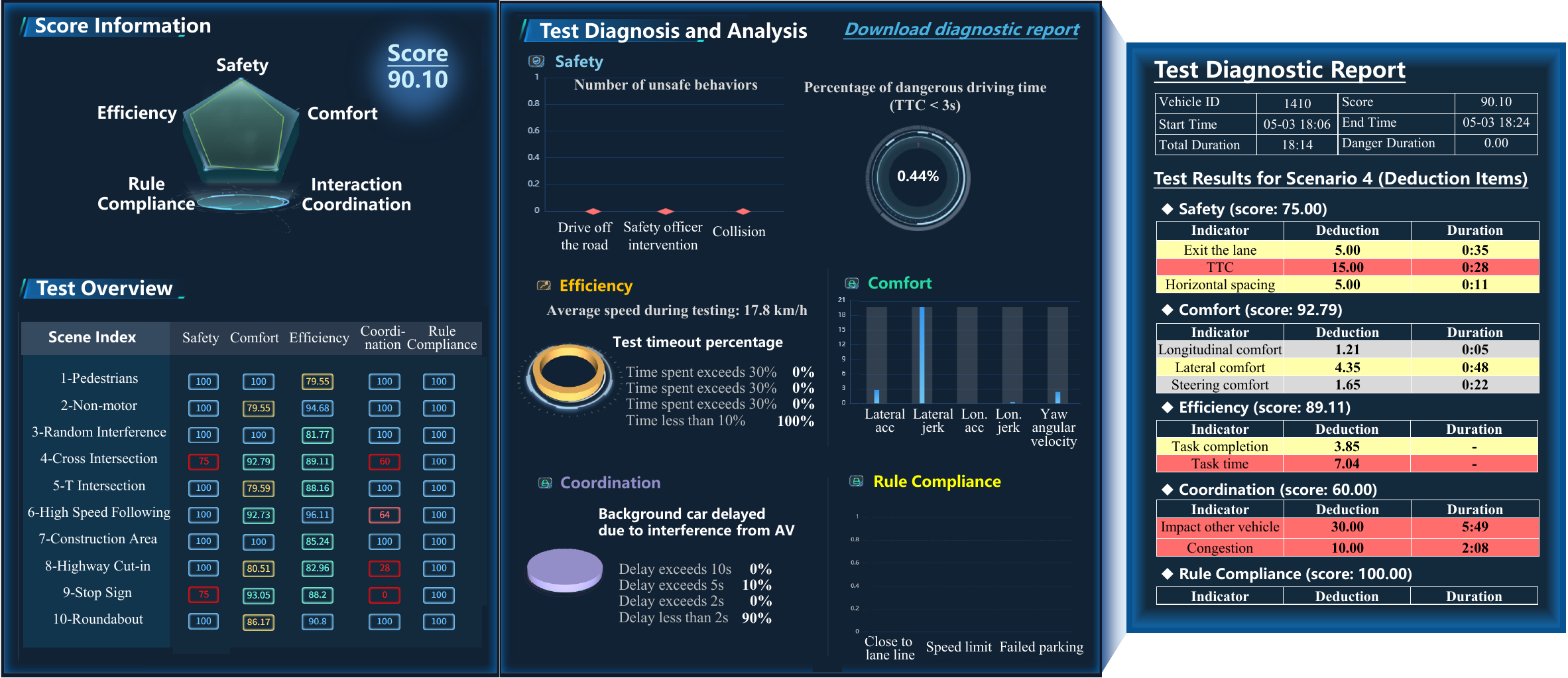}}
  \caption{The five-dimensional intelligence evaluation results of a given AUT, together with its corresponding test diagnostic report.}\label{Analysis}
  \end{center}
\end{figure*}

Fig.~\ref{Analysis} shows the test result analysis and evaluation interface for one autonomous driving algorithm. For each test scenario, there are sub-scores across five dimensions, with key metric analysis results displayed on the right-hand side of the interface. Users can export the \textbf{\textit{Test Diagnostic Report}} to view the specific scores for each low-scoring dimension, guiding targeted improvements to the algorithm. For example, the algorithm tested in Fig.~\ref{Analysis} performs well in rule compliance and comfort, but it received lower scores in safety (TTC), efficiency (task completion time), and coordination, indicating direction of improvement.

\subsection{Platform Credibility Assessment}
A critical prerequisite for leveraging a virtual-physical fusion testing platform is to ensure that its results faithfully represent the real-world performance of the autonomous driving system under test. 
A key prerequisite for utilizing the virtual–physical fusion testing platform is ensuring that its test results accurately reflect the real-world performance of the autonomous driving system under evaluation. Therefore, it is necessary to conduct a systematic assessment of the platform’s credibility. By comparing the real-world road test data with VP-AutoTest’s test results, the platform’s consistency across various scenarios is validated. Furthermore, a multi-dimensional set of metrics is employed for quantitative analysis, comprehensively measuring the fidelity and reliability of the platform.

The methodology unfolds in five stages. First, representative test scenarios are executed in the real world, with both the VUT in autonomous mode and background vehicles operated by human drivers as physical entities. Second, the corresponding fusion test scenarios are constructed within VP-AutoTest. The collected data include time-stamped positions, velocities, and accelerations for VUT and background vehicles. Third, to account for variability between runs, the time-series trajectories are temporally aligned using Dynamic Time Warping (DTW) \cite{senin2008dynamic}. Fourth, considering that directly comparing high-dimensional trajectories is complex, Principal Component Analysis (PCA) \cite{mackiewicz1993principal}is applied to reduce dimensionality while preserving key dynamic behaviors. Finally, the dimensionality-reduced trajectories are quantitatively evaluated using multi-dimensional metrics to provide a comprehensive credibility assessment.

To ensure a robust and comprehensive evaluation, this study selects five different metrics. Each metric is chosen to investigate a specific facet of credibility, collectively covering statistical correlation, error magnitude, non-linear complexity, and frequency-domain characteristics.
\subsubsection{Pearson Correlation Coefficient (PCC) \cite{benesty2009pearson}}
This metric quantifies the strength and direction of the linear relationship between the real-world and fusion test data series. Values closer to +1 indicate a strong positive correlation, meaning that the fusion test successfully reproduces the general behavioral trends of the real system.
\subsubsection{Root Mean Squared Error (RMSE)}
RMSE measures the overall point-by-point deviation between the trajectories from the two test sources. It penalizes large errors more heavily, providing a robust assessment of the fusion test’s accuracy. Lower values indicate smaller discrepancies and higher fidelity.
\subsubsection{Theil's Inequality Coefficient (TIC) \cite{song2013method}}
TIC evaluates the statistical discrepancy between fusion test and real-world data, while also enabling decomposition of the error into meaningful components. Values approaching 0 indicate a close fit, offering insight into where deviations occur and how well the test replicates reality.
\subsubsection{Cross-Fuzzy Entropy (Cross-FuzzyEn) \cite{xie2010cross}}
This non-linear metric quantifies the degree of pattern asynchrony between two time series. It captures differences in complex dynamic behaviors beyond simple point-wise errors. Lower values indicate higher similarity in temporal patterns and coordinated dynamics.
\subsubsection{Cosine Similarity of Power Spectral Density  \cite{youngworth2005overview}}
CS-PSD assesses credibility in the frequency domain by comparing the power spectral density of trajectories from the two test sources. A similarity score near 1 indicates that motion characteristics, such as vibrations and oscillations, are closely matched, reflecting high-fidelity dynamic behavior reproduction.

\begin{figure}[htbp]
  \begin{center}
  \centerline{\includegraphics[width=3.5in]{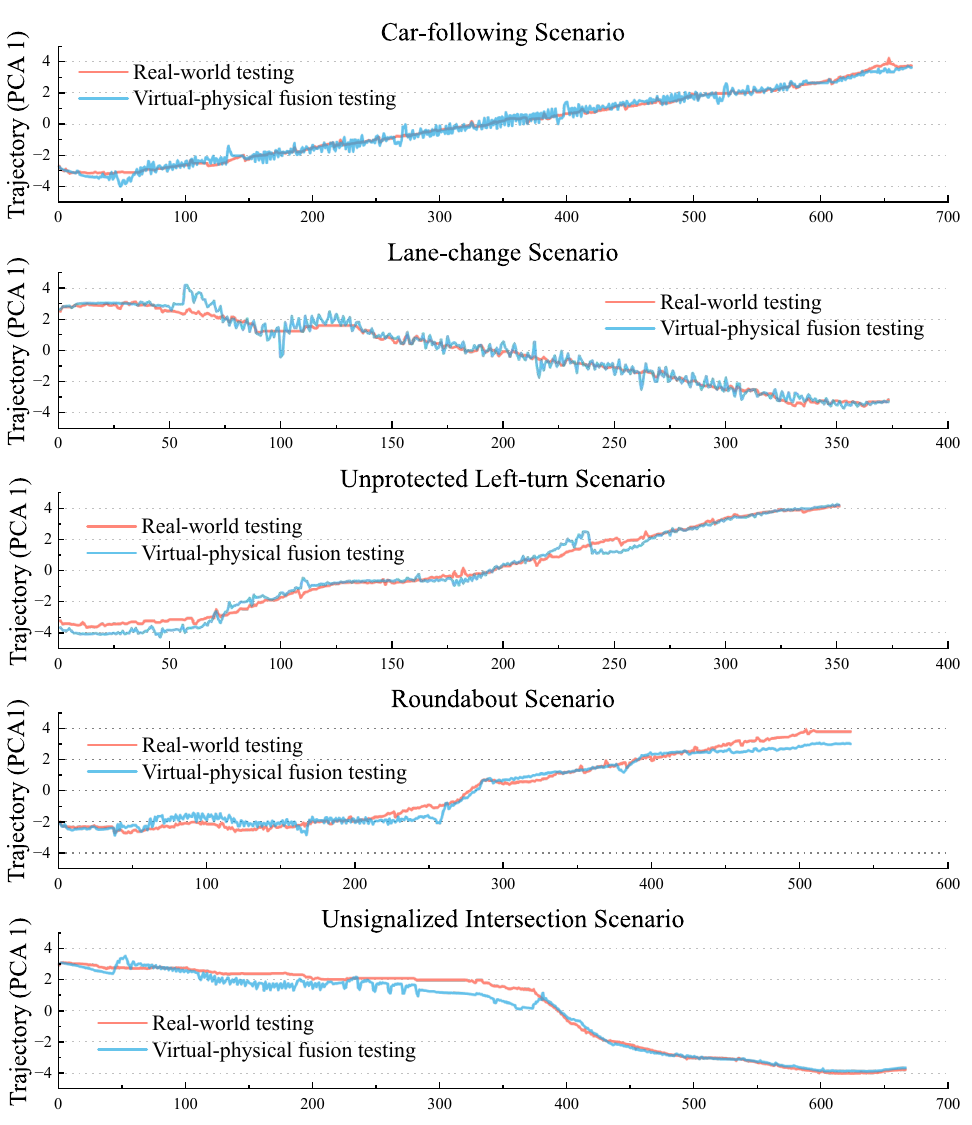}}
  \caption{Trajectories after dimensionality reduction in different scenarios.}\label{AT-traj}
  \end{center}
\end{figure}

The methodology was validated across five representative driving scenarios: car-following, lane-changing, unprotected left turns, roundabouts, and unsignalized intersections. Fig.\ref{AT-traj} presents the first principal component (PCA 1) derived from the collective trajectories of all vehicles in both testing modalities. The results reveal that while the overall traffic dynamics captured by the virtual-physical fusion test are broadly consistent with the real-world, they exhibit more pronounced oscillations. This deviation is primarily attributed to system latency and reduced model fidelity, notably the use of simplified dynamic models for background vehicles.
Overall, the test results demonstrate a high degree of consistency with real-world outcomes, with discrepancies remaining within an acceptable range, thereby validating the platform’s credibility.

Table.~\ref{tab:credib_results} presents the credibility assessment results for the proposed virtual-physical testing platform. Across all evaluated scenarios, the PCC and the CS-PSD consistently approach the ideal value of 1.0. This indicates a high degree of fidelity, demonstrating strong correlation between the virtual-physical and real-world test data in both the time and frequency domains. Regarding error metrics, the RMSE reveals a predictable increase in absolute error as scenario complexity rises, though the deviation remains minimal. This trend is corroborated by the TIC, a scale-invariant metric which also confirms the small relative error between the two testing methodologies. Critically, the Cross-FuzzyEn yields its lowest values in the most complex interactive scenarios. This key finding signifies that the platform accurately reproduces the dynamic patterns and temporal synchrony of the real vehicle's behavior, even with a slight increase in numerical error. Therefore, the virtual-physical testing platform demonstrates high credibility across multiple analytical dimensions, positioning it as a robust and reliable tool for developing and validating autonomous driving algorithms.

\begin{table}[h]
\centering
\caption{Results of credibility assessment metrics in different scenarios}
\label{tab:credib_results}
\setlength{\tabcolsep}{3pt}
\begin{tabular}{lccccc}
\toprule
Scenario & PCC & RMSE & TIC & Cross-FuzzyEn & CS-PSD \\
\midrule
Car-following & 0.993 & 0.247 & 0.059 & 0.143 & 0.997 \\
Lane-changing & 0.985 & 0.393 & 0.092 & 0.269 & 0.982 \\
Unprotected left turn & 0.989 & 0.378 & 0.075 & 0.065 & 0.999 \\
Roundabout & 0.985 & 0.431 & 0.101 & 0.056 & 0.991 \\
Unsignalized intersection & 0.990 & 0.445 & 0.087 & 0.036 & 0.998 \\
\bottomrule
\end{tabular}
\end{table}




\section{Conclusion}

This study presents a comprehensive testing environment for AD capability and a system solution built on a virtual–physical fusion testing platform. The proposed VP-AutoTest integrates physical testing elements, including CAVs, HDVs, cloud-controlled intelligent targets, and RSUs, along with virtual testing elements such as virtual CAVs, remotely operated HDVs, and controllable virtual background traffic flows. It supports a wide range of single-vehicle and multi-vehicle cooperative capability tests, encompassing adversarial testing, parallel deduction, V2V cooperation, and V2I cooperation. Meanwhile, the platform establishes a multi-dimensional comprehensive evaluation system, aiding precise problem diagnosis and optimization. It also possesses system-level credibility evaluation capabilities, providing a significant reference and benchmark for the field of virtual–physical fusion testing.
To date, the platform has successfully supported the inaugural OnSite real-world algorithm challenge, whose summary video of the event can be accessed here \footnote{https://drive.google.com/drive/folders/17cbdzwhSDeSUI5Os3EZ4ro61J8HZRQoe?}.

In the future, the platform will continue to expand scenario coverage and promote the development of region-level coordinated testing capabilities. Concurrently, it will align with the continuously evolving regulations and standards for on-road AD, refining standardized evaluation schemes for multi-functional testing to support vehicle safety certification. Moreover, as end-to-end (E2E) AD algorithms gain prominence in both academia and industry, the platform will increasingly cater to their testing requirements, enabling full-process validation and performance assessment to facilitate efficient testing, verification, and iterative improvement of E2E algorithms. By leveraging high-fidelity data from virtual-physical fusion scenarios, especially rare and high-risk events, an automated and intelligent test–train integration loop can be established. In this loop, adaptive testing identifies algorithm weaknesses, and targeted data-driven training addresses them, thereby accelerating iterative development and enhancing the robustness of AD systems.

\ifCLASSOPTIONcaptionsoff
  \newpage
\fi

\footnotesize
\bibliographystyle{IEEEtranN}
\bibliography{IEEEabrv,Bibliography}

\vfill
\end{document}